\documentclass[10pt,twocolumn,letterpaper]{article}

\usepackage[table]{xcolor}
\usepackage{iccv}
\usepackage{times}
\usepackage{epsfig}
\usepackage{graphicx}
\usepackage{amsmath}
\usepackage{amssymb}

\usepackage[breaklinks=true,bookmarks=false]{hyperref}

\usepackage[capitalize]{cleveref}
\crefname{section}{Sec.}{Secs.}
\Crefname{section}{Section}{Sections}
\Crefname{table}{Table}{Tables}
\crefname{table}{Tab.}{Tabs.}

\usepackage{amsmath,amsfonts,bm}

\def\eqref#1{equation~\ref{#1}}

\def\1{\bm{1}}

\DeclareMathAlphabet{\mathsfit}{\encodingdefault}{\sfdefault}{m}{sl}
\SetMathAlphabet{\mathsfit}{bold}{\encodingdefault}{\sfdefault}{bx}{n}

\newcommand{\R}{\mathbb{R}}

\DeclareMathOperator*{\argmin}{arg\,min}

\usepackage{hyperref}
\usepackage{url}
\usepackage{overpic}
\usepackage[accsupp]{axessibility}

\usepackage{graphicx}
\usepackage{ifthen}

\usepackage{epsfig}
\usepackage{graphicx}
\usepackage{color}
\usepackage{microtype}
\usepackage{algorithm}
\usepackage{algorithmic}
\usepackage{ifthen}
\usepackage{booktabs}
\usepackage{enumitem}
\usepackage{caption} 
\usepackage{makecell}
\usepackage{multirow}
\usepackage{xspace}
\usepackage{wrapfig}
\usepackage{graphicx}
\usepackage{braket}

\usepackage{tikz}
\usepackage{comment}

\usepackage{amsthm}
\usepackage{amsmath,amssymb} 
\usepackage{color}
\usepackage{cite}
\usepackage{colortbl,booktabs}
\usepackage[table]{xcolor}
\usepackage[accsupp]{axessibility}  

\newcommand{\filename}[1]{\url{#1}}
\newcommand{\foldername}[1]{\url{#1}}

\let \bs=\boldsymbol

 \global\long\def\R{\mathbb{R}}

\newtheorem{prop}{\textbf{Proposition}}
\newtheorem{defn}{\textbf{Definition}}

\iccvfinalcopy 

\def\iccvPaperID{10281} 

\ificcvfinal\fi

\begin{document}

\title{Implicit Autoencoder for Point-Cloud Self-Supervised Representation Learning}

\author{Siming Yan\textsuperscript{1}
\hspace{0.05in}
Zhenpei Yang\textsuperscript{1} 
\hspace{0.05in}
Haoxiang Li\textsuperscript{2} 
\hspace{0.05in}
Chen Song\textsuperscript{1} 
\hspace{0.05in}
Li Guan\textsuperscript{2} 
\hspace{0.05in}
Hao Kang\textsuperscript{2} 
\hspace{0.05in}
\vspace{4pt}
\\
Gang Hua\textsuperscript{2} \hspace{0.3in}
Qixing Huang\textsuperscript{1}
\vspace{4pt}
\\
\textsuperscript{1}The University of Texas at Austin \hspace{0.3in} \textsuperscript{2}Wormpex AI Research
}

\maketitle
\ificcvfinal\thispagestyle{empty}\fi

\newcommand{\fix}{\marginpar{FIX}}
\newcommand{\new}{\marginpar{NEW}}

\newcommand{\ouralg}{IAE\xspace}

\renewcommand{\baselinestretch}{0.99}

\begin{abstract}
This paper advocates the use of implicit surface representation in autoencoder-based self-supervised 3D representation learning. The most popular and accessible 3D representation, i.e., point clouds, involves discrete samples of the underlying continuous 3D surface. This discretization process introduces sampling variations on the 3D shape, making it challenging to develop transferable knowledge of the true 3D geometry. In the standard autoencoding paradigm, the encoder is compelled to encode not only the 3D geometry but also information on the specific discrete sampling of the 3D shape into the latent code. This is because the point cloud reconstructed by the decoder is considered unacceptable unless there is a perfect mapping between the original and the reconstructed point clouds. This paper introduces the Implicit AutoEncoder (IAE), a simple yet effective method that addresses the sampling variation issue by replacing the commonly-used point-cloud decoder with an implicit decoder. The implicit decoder reconstructs a continuous representation of the 3D shape, independent of the imperfections in the discrete samples. Extensive experiments demonstrate that the proposed IAE achieves state-of-the-art performance across various self-supervised learning benchmarks. Our code is available at \href{https://github.com/SimingYan/IAE}{https://github.com/SimingYan/IAE}.

\end{abstract}

\section{Introduction}
\label{Section:Introduction}

The rapid advancement and growing accessibility of commodity scanning devices have made it easier to capture vast amounts of 3D data represented by point clouds. The success of point-based deep networks~\cite{qi2017pointnet, qi2017pointnet++} additionally enables many 3D vision applications to exploit this natural and flexible representation. Exciting results have emerged in recent years, ranging from object-level understanding, including shape classification~\cite{chang2015shapenet} and part segmentation~\cite{yi2016scalable, sharma2020parsenet, yan2021hpnet}, to scene-level understanding, such as 3D object detection~\cite{dai2017scannet, song2015sun, geiger2012we} and 3D semantic segmentation~\cite{armeni20163d}. 
\begin{figure}[t]
    \centering
    \includegraphics[width=0.8\columnwidth]{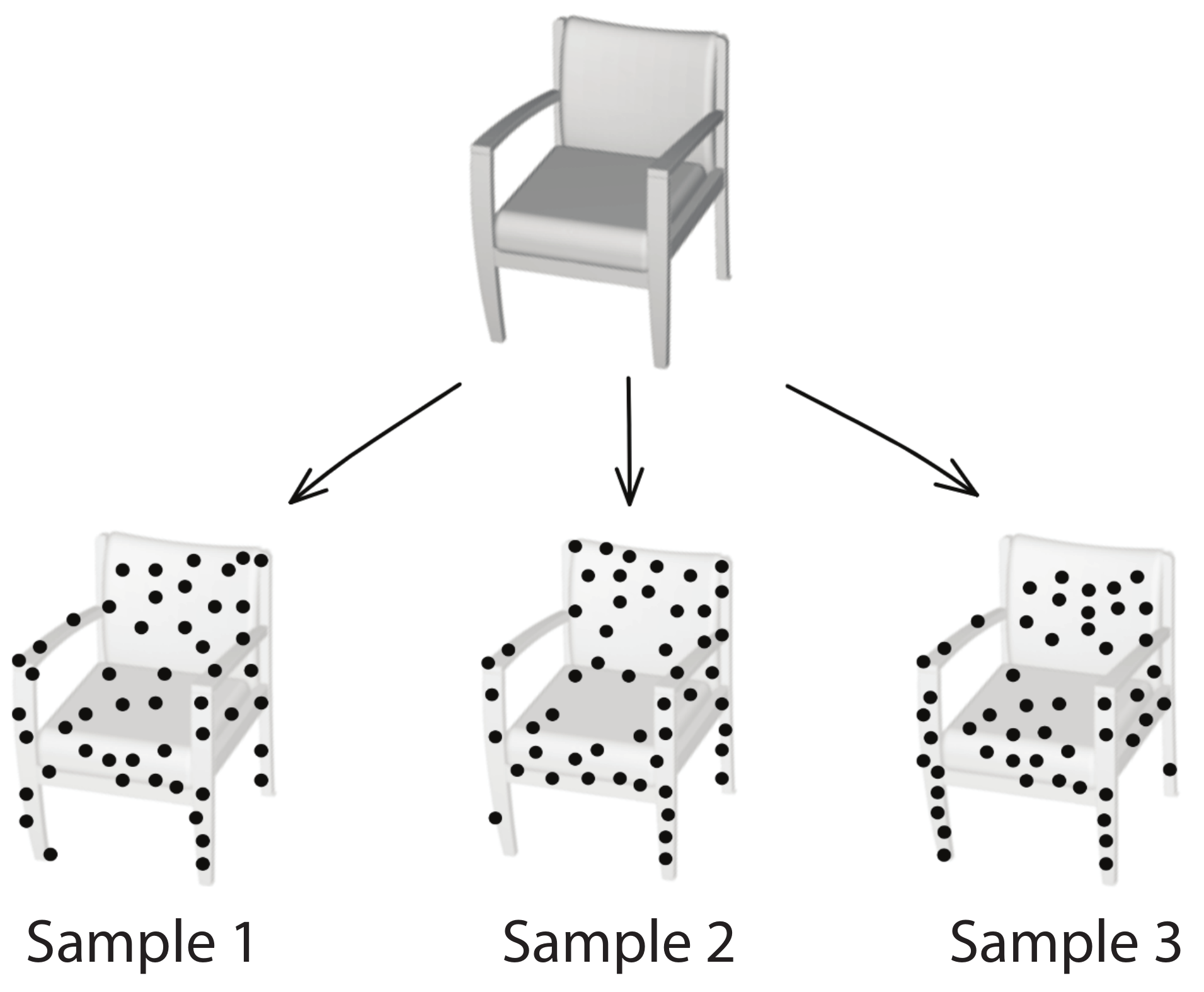}
    \caption{\small{\textbf{The Sampling Variation Problem}. Given a continuous 3D shape, there are infinitely many ways to sample a point cloud. The proposed Implicit AutoEncoder (IAE) learns a latent representation of the true 3D geometry independent of the specific discrete sampling process. By alleviating the sampling variation problem, IAE improves existing point-cloud self-supervised representation learning methods in various downstream tasks.}}
    \label{fig:teaser}
    \vspace{-10pt}
\end{figure}

Annotating point-cloud datasets is a highly labor-intensive task due to the challenges involved in designing 3D interfaces and visualizing point clouds. As a result, researchers are motivated to explore self-supervised representation learning paradigms. The fundamental concept is to pre-train deep neural networks on large unlabeled datasets and fine-tune them on smaller datasets annotated based on specific task requirements. A carefully designed self-supervised representation learning paradigm effectively initializes the network weights, enabling fine-tuning on downstream tasks ({\em e.g.}, classification and segmentation) to avoid weak local minima and achieve improved stability~\cite{erhan2010does}.

Research in self-supervised representation learning predates the history of 3D computer vision. Notably, significant effort has been dedicated to developing self-supervised learning methods for 2D images~\cite{pathak2016context, wu2018unsupervised, doersch2015unsupervised, masci2011stacked, chen2020simple, zhuang2019self, zhuang2021unsupervised}, with autoencoders being one of the most popular tools~\cite{pathak2016context,bengio2013representation,tschannen2018recent,vincent2008extracting,he2021masked}. An autoencoder-based self-supervised representation learning pipeline comprises an encoder that transforms the input into a latent code and a decoder that expands the latent code to reconstruct the input. Since the latent code has a much lower dimension than the input, the encoder is encouraged to summarize the input by condensed latent features with the help of the reconstruction loss.

There is a growing interest in developing self-supervised representation learning methods for point clouds by drawing inspiration from image-based autoencoders. For example, Yang et al.~\cite{yang2018foldingnet} propose a novel folding-based decoder, and Wang et al.~\cite{wang2020unsupervised} develop a denoising autoencoder based on the standard point-cloud completion model. However, to the best of our knowledge, prior works in point-cloud self-supervised representation learning have relied on the same design principles as image-based methods, where \textbf{both the encoder and decoder represent the 3D geometry in the same format} ({\em i.e.}, point clouds).

Point clouds are noisy, discretized, and unstructured representations of 3D geometry. As shown in Figure~\ref{fig:teaser}, a 3D shape can be represented by many different point clouds, all of which are valid representations of the same object. Different point cloud samples are subject to different noises, which are induced from various sources such as the intrinsic noises from the sensors and the interference from the environment. The unordered and noisy nature distinguishes point clouds from conventional structured data, such as pixel images defined on rectangular grids.  When training a point-cloud autoencoder, the encoder is forced to capture \textit{sampling variations}, limiting the model's ability to extract valuable information about the true 3D geometry.

This paper, for the first time, formalizes the concept of sampling variations and proposes to combine the implicit surface representation with point-cloud self-supervised representation learning. Specifically, we introduce an asymmetric point-cloud autoencoder scheme, where the encoder takes a point cloud as input, and the decoder uses the implicit function as the output 3D representation. The rest of the paper refers to this design as the Implicit AutoEncoder (IAE). IAE enjoys many advantages over existing methods. First, reconstruction under the implicit representation discourages latent space learning from being distracted by the imperfections brought by sampling variations and encourages the encoder to capture generalizable features from the true 3D geometry. Second, the reconstruction loss defined on implicit functions bypasses the expensive and unstable data association operation commonly used in point-based reconstruction losses such as the Earth Mover Distance~\cite{rubner2000earth} and the Chamfer Distance~\cite{fan2017point}. The added efficiency allows IAE to process up to 40k input points with a single Tesla V100 GPU, making it possible to capture very fine geometric details when existing methods can only work with sparse data with approximately 1k points.

To demonstrate the effectiveness of IAE, we conduct experiments to verify that the learned representation from our pre-trained model can adapt to various downstream tasks, both at the object level and the scene level, including shape classification, linear evaluation, object detection, and indoor semantic segmentation. IAE consistently outperforms state-of-the-art methods in all settings. 
Specifically, under the best setting, IAE achieves 88.2\% / 94.3\% classification accuracy on ScanObjectNN~\cite{uy-scanobjectnn-iccv19} / ModelNet40~\cite{wu20153d} respectively.
IAE is also the first to support the autoencoding paradigm in scene-level pre-training for the application to various scene-level downstream tasks. For example, compared to training from scratch, IAE achieves +0.7\% and +1.1\% absolute improvements in object detection quality evaluated by mAP@0.5 on ScanNet~\cite{dai2017scannet} and SUN RGB-D~\cite{song2015sun}, respectively.

We summarize the contributions of our paper as follows:

\begin{enumerate}[leftmargin=*]\setlength\itemsep{0mm}
    \item[-] We propose an asymmetric point-cloud autoencoder called Implicit AutoEncoder (IAE). The IAE takes a point cloud as input and uses an implicit function as the output 3D representation. We combine the implicit surface representation with point-cloud self-supervised representation learning for the first time.
    \item[-] We formalize the concept of
sampling variations in point clouds and demonstrate that IAE is more effective at capturing generalizable features from true 3D geometry than standard point-cloud autoencoders.
    \item[-] We conduct experiments to demonstrate the effectiveness of IAE on various downstream tasks, including shape classification, object detection, and indoor semantic segmentation. IAE consistently outperforms state-of-the-art methods in all settings.
    
\end{enumerate}

\section{Related Work}
\noindent\textbf{Point-Cloud Self-Supervised Representation Learning}. The unordered nature of 3D point clouds poses a unique challenge to representation learning, which is known to be effective on conventional structured data representations like images. Several point-cloud self-supervised methods have been developed to learn representations from individual 3D objects~\cite{wang2020unsupervised, sauder2019self, yang2018foldingnet, poursaeed2020self, achlioptas2018learning, hassani2019unsupervised, li2018so, chen2021shape, yu2021point, yan20233d}, with the synthetic ShpapeNet~\cite{chang2015shapenet} being the most popular dataset used in pre-training. However, existing methods exhibit limited generalizability to real scene-level datasets due to a significant domain gap~\cite{xie2020pointcontrast, yan2023learning}.

The remarkable success achieved in 2D has sparked a growing interest in extending the contrastive learning paradigm to self-supervised representation learning on point clouds~\cite{xie2020pointcontrast, rao2021randomrooms, zhang2021self}. These approaches rely on the contrastive loss to separate positive and negative instance pairs in the latent space, necessitating meticulous design efforts to define similar and dissimilar examples. They also consume a substantial amount of computational resources, particularly when working with large datasets. While 3D contrastive learning is a popular field of research, this paper addresses the sampling variation problem under the simple autoencoder paradigm discussed below.

\noindent\textbf{Point-Cloud Autoencoders}. Several existing works exploit the autoencoder paradigm for point-cloud self-supervised representation learning. They simultaneously train an encoder and a decoder, with the encoder mapping the input point cloud to a latent code, and the decode reconstructing another point cloud from the latent code~\cite{yang2018foldingnet, wang2020unsupervised, pang2022masked, zhang2022point}. While these approaches have demonstrated promising performance improvements, the sampling variation problem remains a significant challenge when the decoder output is represented as point clouds. This is because the network must encode not only the geometry but also the imperfections of the specific point cloud sampling into the latent code. Additionally, training a decoder under a point-based network architecture is not as straightforward as training a regular convolutional or fully-connected model, which limits the quality of the output. This paper proposes the use of implicit representation as the decoder output in the autoencoder-based self-supervised representation learning paradigm. As we will see, the implicit decoder alleviates the sampling variation issue and allows for a more efficient pipeline than existing approaches. Closely related to our design, ParAE~\cite{eckart2021self} learns the 3D data distribution through discrete generative models. While ParAE builds the supervision upon a point-wise partitioning matrix, the proposed IAE utilizes implicit representations that are conceptually more relevant to the 3D geometry and easier to train.

\noindent\textbf{Implicit Representations}. A 3D shape can be represented implicitly by a deep network that maps 3D coordinates to signed distances~\cite{michalkiewicz2019implicit, park2019deepsdf} or occupancy grids~\cite{chen2019learning, mescheder2019occupancy, peng2020convolutional}. Unlike explicit representations such as point clouds, voxels, or triangle meshes, implicit representations offer a continuous shape parameterization and do not suffer from discretization errors. Implicit representations have been successfully applied in various 3D tasks, including 3D reconstruction from images~\cite{liu2020dist, liu2019learning, niemeyer2020differentiable}, primitive-based 3D reconstruction~\cite{genova2020local, genova2019learning, paschalidou2020learning}, 4D reconstruction~\cite{niemeyer2019occupancy}, and continuous texture representation~\cite{oechsle2019texture}. However, to the best of our knowledge, no existing research has investigated the integration of implicit representations into the self-supervised representation learning paradigm, making IAE the pioneer in the field.

\section{Method}
\label{sec:method}

\subsection{Problem Setup}
Assume we have access to a large dataset of 3D point clouds $\mathbf{D} = \{\mathcal{P} | \mathcal{P} \in \mathbb{R}^{n \times 3}\}$ without any annotations. The goal of autoencoder-based point-cloud self-supervised learning is to train an encoder network $f_{\Theta}$ and a decoder network $g_{\Phi}$, where the encoder maps the point cloud to an m-dimensional latent code:
\begin{equation}
\small
    f_{\Theta}: \mathbb{R}^{n \times 3} \rightarrow \mathbb{R}^{m},  m \ll n
\end{equation}
and the decoder maps the latent back to the 3D geometry:
\begin{equation}
\small
    g_{\Phi}: \mathbb{R}^{m} \rightarrow \mathbb{S}
\end{equation}
Here, $\mathbb{S}$ refers to the space of 3D shapes under the representation of choice. In practice, a popular choice is to use a sub-sampled version of the point cloud as the output shape representation: $\mathbb{S} = \mathbb{R}^{n' \times 3}$, where $n' \leq n$. The parameters $\Theta$ and $\Phi$ are jointly trained by minimizing the distance metric $d$ between the input point cloud and the reconstructed shape:
\begin{equation}
\small
    \Theta^*, \Phi^* = \argmin_{\Theta, \Phi} d((g_{\Phi} \circ f_{\Theta})(\mathcal{P}), \mathcal{P})
\end{equation}
After training the autoencoder, the encoder $f_{\Theta^*}$ is fine-tuned on a small dataset with task-specific annotations ({\em e.g.}, bounding boxes and segmentation labels).

\subsection{Challenges}
\label{sec:challenge}
\noindent\textbf{Sampling Variations}. As shown in Figure~\ref{fig:teaser}, there are infinitely many valid point clouds representing the same 3D shape. This is because point clouds are discrete samples of the continuous 3D geometry. Due to the finite sampling size and noisy conditions, each point cloud contains a unique set of imperfections from the sampling process. Ideally, an autoencoder is expected to encode the information just about the true 3D geometry into the latent code. In practice, the encoder is compelled to capture the imperfections in the latent code since the decoder output is represented as another point cloud with a distance metric in place enforcing the two discrete samples to be identical. Sampling variation causes distractions to the autoencoder learning and leads to suboptimal performance in downstream tasks. 

\noindent\textbf{Efficient Distance Calculation}. The Earth Mover Distance (EMD)~\cite{rubner2000earth} in Eq.~\ref{eq:emd} and the Chamfer Distance (CD)~\cite{fan2017point} in Eq.~\ref{eq:cd} are commonly used metrics to calculate the distance between the input and the reconstructed point clouds. For each point $\mathbf{x}$ in point cloud $\mathcal{P}$, EMD and CD find the nearest point in the other cloud $\hat{\mathcal{P}}$ for distance calculation. The correspondence discovery operation makes both EMD and CD expensive metrics to use, especially when the point clouds are dense. 

\begin{equation}
\small
    d_\text{EMD}(\mathcal{P}, \hat{\mathcal{P}}) =  \min_{\phi: \mathcal{P} \rightarrow \hat{\mathcal{P}}}\sum_{\mathbf{x} \in \mathcal{P}}||\mathbf{x} - \phi(\mathbf{x})||_2
    \label{eq:emd}
\end{equation}
\vspace{-4pt}
\begin{equation}
    \small
    d_\text{CD}(\mathcal{P}, \hat{\mathcal{P}}) = \sum_{\hat{\mathbf{x}} \in \hat{\mathcal{P}}} \min_{\mathbf{x} \in \mathcal{P}} ||\hat{\mathbf{x}} - \mathbf{x}||_2 +  \sum_{\mathbf{x} \in \mathcal{P}}\min_{\hat{\mathbf{x}} \in \hat{\mathcal{P}}} ||\mathbf{x} - \hat{\mathbf{x}}||_2
    \label{eq:cd}
\end{equation}

\begin{figure*}[h]
\centering
\includegraphics[width=.95\textwidth]{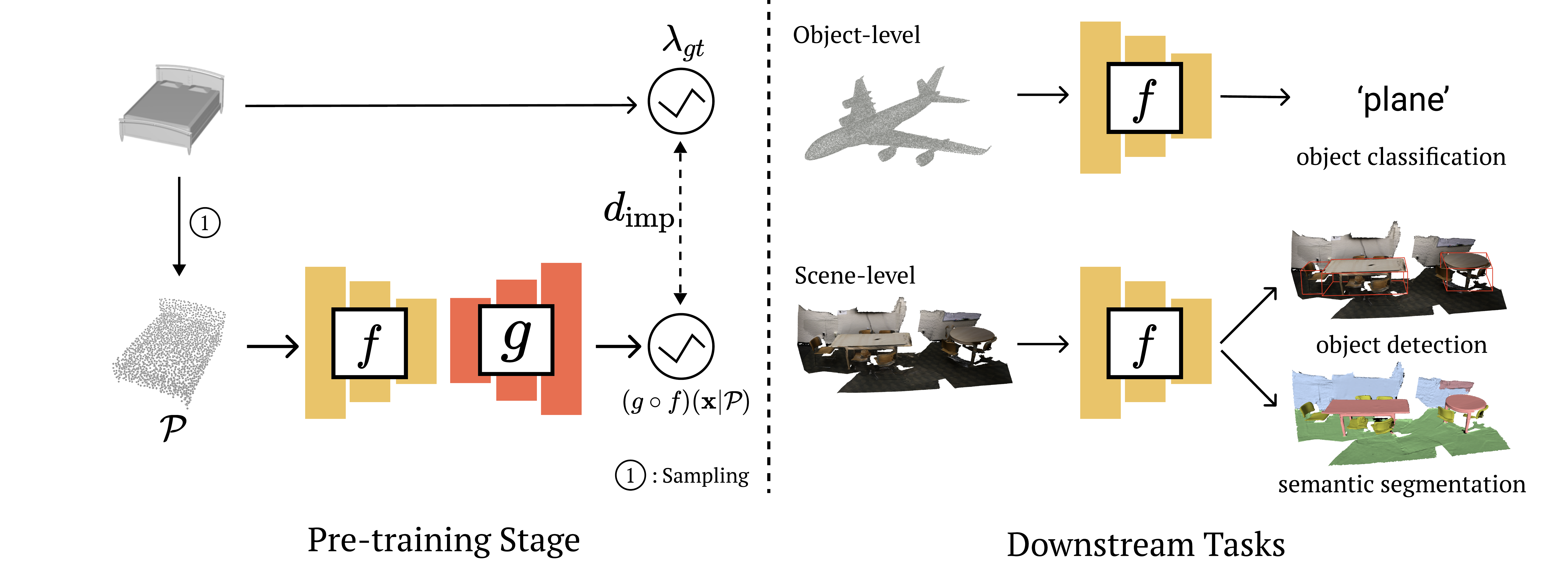}
\caption{\small{\textbf{The Implicit AutoEncoder.} In the \textbf{pre-training stage}, we jointly train an encoder $f_{\Theta}$ and a decoder $g_{\Phi}$ to reconstruct the underlying 3D shape from the input\protect\footnotemark. While the encoder takes the point cloud as input, the output of the decoder is represented by an implicit function. The implicit function maps a 3D query point $\mathbf{x} = (x, y, z)$ to the signed~\cite{michalkiewicz2019implicit, park2019deepsdf}, unsigned~\cite{chibane2020neural}, or occupancy~\cite{chen2019learning, mescheder2019occupancy, peng2020convolutional} representation of the shape. After pre-training, we fine-tune the encoder $f_{\Theta}$ on different \textbf{downstream tasks}. At the object level, we evaluate IAE on object classification. At the scene level, we evaluate IAE on object detection and semantic segmentation.}}
\label{fig:main_figure}
\vspace{-0.1in}
\end{figure*}

\noindent\textbf{Data Density}. Many large-scale open 3D datasets provide very dense point clouds. For example, the popular ShapeNet dataset~\cite{chang2015shapenet} contains at least 50k points per shape. Due to the inefficiency discussed above, existing explicit autoencoders lack the ability to exploit such density and are limited to a sparely sub-sampled version of the original dense point cloud. Sub-sampling introduces information loss and additional sampling variations. If it \textit{could} use dense point clouds during training, an explicit autoencoder \textit{would} potentially be as good as the proposed IAE. The problem, however, is that it is non-trivial to use dense data in explicit autoencoders. A naive adoption of the EMD~\cite{rubner2000earth} and CD~\cite{fan2017point} losses leads to an unaffordable computation cost. We show more analysis in Section~\ref{sec:analysis}.

\footnotetext{For simplification, we denote $f$ as $f_{\Theta}$, $g$ as $g_{\Phi}$, and do not show the query point $\mathbf{x}$ in Figure~\ref{fig:main_figure}.} 

\subsection{The Implicit AutoEncoder (IAE)}
\noindent\textbf{Design}.
As illustrated by Figure~\ref{fig:main_figure}, we propose the Implicit AutoEncoder (IAE), which elegantly addresses the challenges discussed above. In IAE, while the encoder still takes a point cloud as input, the decoder outputs an implicit function: $\mathbb{S} = \{\lambda | \lambda(\mathbf{x}) \in \mathbb{R}, \forall \mathbf{x} \in \mathbb{R}^3 \}$. Let $\lambda_\text{gt}$ be the ``ground-truth'' implicit function derived analytically from the input shape, the training objective of IAE is to minimize the distance between two implicit surfaces through the distance metric $d_\text{imp}$:
\begin{equation}
\small
    \Theta^*, \Phi^* = \argmin_{\Theta, \Phi} d_\text{imp}((g_{\Phi}\circ f_{\Theta})(\mathbf{x}|\mathcal{P}), \lambda_\text{gt}(\mathbf{x}))
    \label{eq:imp}
\end{equation}
Our experiments use three popular variants of the implicit function, including the signed distance function~\cite{michalkiewicz2019implicit, park2019deepsdf}, the unsigned distance function~\cite{chibane2020neural}, and the occupancy grid~\cite{chen2019learning, mescheder2019occupancy, peng2020convolutional}. When the output shape is represented by either the signed (SDF) or the unsigned (UDF) distance function, $d_\text{imp}$ is implemented as the mean $\mathcal{L}_1$ distance between the ground-truth and reconstructed distance functions on $N$ uniformly sampled points inside the 3D bounding box:

\begin{equation}
\small
    d_\text{sdf} = \frac{1}{N}\sum_{\mathbf{x} \in X} |(g_{\Phi}\circ f_{\Theta})(\mathbf{x}|\mathcal{P}) - s(\mathbf{x})|
    \label{eq:sdf}
\end{equation}
\begin{equation}
\small
    d_\text{udf} = \frac{1}{N}\sum_{\mathbf{x} \in X} ||(g_{\Phi}\circ f_{\Theta})(\mathbf{x}|\mathcal{P})| - u(\mathbf{x})|
    \label{eq:sdf}
\end{equation}
where $s$ denotes SDF value of sampled point $\mathbf{x}$, and $u$ denotes UDF value of $\mathbf{x}$.

On the other hand, when the output shape is represented by the occupancy grid, $d_\text{imp}$ becomes the cross-entropy loss:

\begin{equation}
\small
    d_\text{occ} = \frac{1}{N} \sum_{\mathbf{x} \in X} CE((g_{\Phi}\circ f_{\Theta})(\mathbf{x}|\mathcal{P}), o(\mathbf{x}))
    \label{eq:occ}
\end{equation}
where $CE$ denotes cross entropy, $o$ denotes occupancy.

\section{Experiments}

\begin{table*}[h]
    \centering
    \setlength\tabcolsep{11pt}
    \small
    \captionof{table}
    {\small{\textbf{Classification results on ScanObjectNN and ModelNet40 datasets.} The model parameters number (\#Params), GFLOPS, and overall accuracy (\%) are reported.} The FULL section presents the results of fine-tuning the pre-trained models. The LINEAR section reports the results of training a linear SVM. To establish a fair comparison, we present the performance of IAE using two distinct architectures: DGCNN and Transformer-like Point-M2AE. These architectures are commonly adopted in prior methods. The methods within the same block share the same architecture. `w/o mesh' indicates that pre-training does not involve mesh data.}
    \begin{tabular}{lccccccc}
        \toprule
        \multirow{2}{*}{Method} & \multirow{2}{*}{\#Params(M)} & \multirow{2}{*}{GFLOPS} & \multicolumn{3}{c}{ScanObjectNN} & \multicolumn{2}{c}{ModelNet40} \\ 
     \cmidrule(lr){4-6}   \cmidrule(lr){7-8}   & & & OBJ\_BG & OBJ\_ONLY & PB\_T50\_RS & 1k P & 8k P \\
        \midrule
        \multicolumn{8}{c}{\textit{Supervised Learning Only}} \\
        \midrule
         PointNet~\cite{qi2017pointnet++} & 3.5 & 0.5 & 73.3 & 79.2 & 68.0 & 89.2 & 90.8 \\
        PointNet++~\cite{qi2017pointnet} & 1.5 & 1.7 & 82.3 & 84.3 & 77.9 & 90.7 & 91.9 \\
        PointCNN~\cite{li2018pointcnn} & 0.6 & - & 86.1 & 85.5 & 78.5 & 92.2 & - \\
        KPConv~\cite{thomas2019kpconv} & - & - & 86.4 & 84.1 & 80.2 & 92.9 & - \\
        GBNet~\cite{qiu2021geometric} & 8.8 & - & - & - & 81.0 & 93.8 & -\\
        PointMLP~\cite{ma2022rethinking} & 12.6 & 31.4 & - & - & 85.7 & 94.1 & -\\
        PointNeXt~\cite{qian2022pointnext} & 1.4 & 3.6 & - & - & 88.1 & 93.2 & -\\
        \midrule
        \multicolumn{8}{c}{\textit{with Self-Supervised Representation Learning} (FULL)} \\
        \midrule
        DGCNN~\cite{wang2019dynamic} & 1.8 & 2.4 & 82.8 & 86.2 & 78.1 & 92.9 & 93.1 \\
        JigSaw~\cite{sauder2019self} & 1.8 & 2.4 & - & - & 83.5 & 92.6 & - \\
        STRL~\cite{huang2021spatio} & 1.8 & 2.4 & - & - & - & 93.1 & -\\
        OcCo~\cite{wang2021unsupervised} & 1.8 & 2.4 & 88.2 & 87.5 & 84.3 & 93.0 & -\\
         IAE (DGCNN) & 1.8 & 2.4 & \textbf{90.2} & \textbf{89.0} & \textbf{85.6} & \textbf{94.2} & \textbf{94.2} \\
        \midrule
        Transformer~\cite{yu2021point} & 22.1 & 4.8 & 83.0 & 84.1 & 79.1 & 91.4 & 91.8\\
        Point-BERT~\cite{yu2021point} & 22.1 & 4.8 & 89.3 & 88.1 & 84.3 & 93.2 & 93.8\\
        Point-MAE~\cite{pang2022masked} & 22.1 & 4.8 & 90.0 & 88.3 & 85.2 & 93.8 & 94.0 \\
        Point-M2AE~\cite{zhang2022point} & 15.3 & 3.6 & 91.2 & 88.8 & 86.4 & 94.0 & - \\      
        IAE (M2AE) w/o mesh & 15.3 & 3.6 & 92.3 & 91.2 & 88.0 & 94.1 & 94.2 \\
        IAE (M2AE) & 15.3 & 3.6 & \textbf{92.5} & \textbf{91.6} & \textbf{88.2} & \textbf{94.2} &\textbf{ 94.3} \\
        \midrule
        \multicolumn{8}{c}{\textit{with Self-Supervised Representation Learning} (LINEAR)} \\
        \midrule
        JigSaw~\cite{sauder2019self} & 1.8 & 2.4 & - & - & 59.5 & 84.1 & - \\
        STRL~\cite{huang2021spatio} & 1.8 & 2.4 & - & - & 77.9 & 90.9 & -\\
        OcCo~\cite{wang2021unsupervised} & 1.8 & 2.4 & - & - & 78.3 & 89.7 & -\\
        CrossPoint~\cite{afham2022crosspoint} & 1.8 & 2.4 & - & - & 81.7 & 91.2 & -\\
        IAE (DGCNN) & 1.8 & 2.4 & - & - & \textbf{83.0} & \textbf{92.1} & -\\
        \midrule
        Point-M2AE~\cite{zhang2022point} & 15.3 & 3.6 & - & - & 83.1 & 92.9 & - \\
        IAE (M2AE) w/o mesh & 15.3 & 3.6 & - & - & 84.3 & 93.1 & - \\
        IAE (M2AE) & 15.3 & 3.6 & - & - & \textbf{84.4} & \textbf{93.2} & - \\
        \bottomrule
    \end{tabular}
    \label{tab:cls}
\vspace{-3mm}
\end{table*}
\subsection{Pre-Training Datasets}
\label{sec:setup}
\noindent\textbf{ShapeNet}~\cite{chang2015shapenet} is a synthetic dataset commonly used in 3D vision, consisting of 57,748 shapes from 55 object categories. In addition to point clouds, ShapeNet offers high-quality water-tight triangular meshes as a more accurate description of 3D geometry. Following~\cite{park2019deepsdf,mescheder2019occupancy}, we generate ground-truth signed distance functions and occupancy grids from the provided mesh data. While ShapeNet is able to provide such high-quality 3D geometric information thanks to its synthetic nature, water-tight meshes are rarely available in real-world settings. To assess the effectiveness of IAE under more realistic conditions, we further use the nearest neighbor strategy~\cite{peterson2009k} to obtain unsigned distance functions from the point clouds in ShapeNet (50k points). 

\noindent\textbf{ScanNet}~\cite{dai2017scannet} is a real-world dataset with over 1,500 point-cloud scans. Unlike ShapeNet~\cite{chang2015shapenet}, which focuses on individual objects, ScanNet collects indoor scene data with a significantly larger perceptive range. We apply the sliding window strategy and crop each instance into $d \times d \times d$ cubes, where $d=3.0$ m. Following Qi et al.~\cite{qi2019deep}, we divide the dataset into training (8k point clouds) and validation (2.6k point clouds) splits. We randomly sub-sample 10,000 points from each point cloud as the input to the encoder. Due to the lack of high-quality water-tight meshes, we use the nearest neighbor strategy~\cite{peterson2009k} to calculate ground-truth unsigned distance functions. We refer interested readers to the supplementary material for implementation details. Importantly, we point out that IAE is the first method that supports the autoencoding paradigm in real-world scene-level training. As observed by Xie et al.~\cite{xie2020pointcontrast}, pre-training on synthetic datasets exhibits limited generalizability to real-world tasks. Real-world scene-level pre-training is extremely challenging due to noises, incomplete scans, and the complication of indoor scene contents. As a result, existing self-supervised methods fail to achieve any noticeable performance improvement~\cite{sauder2019self, poursaeed2020self, wang2020unsupervised}.

\begin{table*}[t]
\small
\begin{minipage}[t]{.55\linewidth}
    \centering
   \setlength\tabcolsep{10pt}
    \begin{tabular}{lcccc}
        \toprule
     Method  & \multicolumn{2}{c}{ScanNet} & \multicolumn{2}{c}{SUN RGB-D} \\
        \cmidrule(lr){2-3}   \cmidrule(lr){4-5}  & $\text{AP}_{25}$ & $\text{AP}_{50}$ & $\text{AP}_{25}$ & $\text{AP}_{50}$ \\
        \midrule
        VoteNet~\cite{qi2019deep}  &  58.6 & 33.5 & 57.7 & 32.9 \\
        3D-MPA~\cite{engelmann20203d} & 64.2 & 49.2 & - & - \\
        MLCVNet~\cite{xie2020mlcvnet} & 64.5 & 41.4 & 59.8 & - \\
        BRNet~\cite{cheng2021back} & 66.1 & 50.9 & 61.1 & 43.7 \\
        3DETR~\cite{misra2021end} & 65.0 & 47.0 & 59.1 & 32.7 \\
        GroupFree~\cite{liu2021group} & 69.1 & 52.8 & 63.0 & 45.2 \\
        FCAF3D~\cite{rukhovich2021fcaf3d} & 71.5 & 57.3 & 64.2 & 48.9 \\
        CAGroup3D~\cite{wang2022cagroup3d} & 75.1 & 61.3 & 66.8 & 50.2 \\
        \midrule
        STRL~\cite{huang2021spatio} & 59.5 & 38.4 & 58.2 & 35.0 \\
        RandomRooms~\cite{rao2021randomrooms} & 61.3 & 36.2 & 59.2 & 35.4 \\
        PointContrast~\cite{xie2020pointcontrast} & 59.2 & 38.0 & 57.5 & 34.8 \\
        DepthContrast\protect\footnotemark~\cite{zhang2021self} & 62.1 & 39.1 & 60.4 & 35.4 \\
        Point-M2AE~\cite{zhang2022point} & 66.3 & 48.3 & - & - \\
        IAE (VoteNet) & 61.5 & 39.8 & 60.4 & 36.0 \\
        IAE (CAGroup3D) & \textbf{76.1} & \textbf{62.0} & \textbf{67.6} & \textbf{51.3}\\
        \bottomrule
    \end{tabular}
    \caption{\small{\textbf{3D Object Detection Results.} We fine-tune our pre-trained model on ScanNetV2~\cite{dai2017scannet} and SUN-RGBD~\cite{song2015sun} validation sets using VoteNet~\cite{qi2019deep} and CAGroup3D~\cite{wang2022cagroup3d}. We show mean Average Precision~(mAP) across all semantic classes with 3D IoU thresholds of 0.25 and 0.5. Methods in the second section denote self-supervised methods.}}
    \label{tab:detection}
\end{minipage}\hspace{.02\linewidth}
\begin{minipage}[t]{.4\linewidth}
 \setlength\tabcolsep{15pt}
    \begin{tabular}{lcc}
        \toprule
        Method & \multicolumn{2}{c}{S3DIS 6-Fold} \\
        \cmidrule(lr){2-3} & OA & mIoU \\
        \midrule
        PointNet~\cite{qi2017pointnet} & 78.5 & 47.6 \\
        PointCNN~\cite{li2018pointcnn} & 88.1 & 65.4 \\
        DGCNN~\cite{wang2019dynamic} & 84.1 & 56.1 \\
        DeepGCN~\cite{li2021deepgcns} & 85.9 & 60.0 \\
        KPConv~\cite{thomas2019kpconv} & - & 70.6 \\
        RandLA-Net~\cite{hu2020randla} & 88.0 & 70.0 \\
        BAAF-Net~\cite{qiu2021semantic} & 88.0 & 70.0 \\
        PointTransformer~\cite{zhao2021point} & 90.2 & 73.5 \\
        CBL~\cite{tang2022contrastive} & 89.6 & 73.1 \\
        PointNeXt~\cite{qian2022pointnext} & 90.3 & 74.9 \\
        \midrule
        Jigsaw~\cite{sauder2019self} & 84.4 & 56.6 \\
        OcCo~\cite{wang2020unsupervised} & 85.1 & 58.5 \\
        STRL~\cite{huang2021spatio} & 84.2 & 57.1 \\ 
        IAE (DGCNN) & 85.9 & 60.7 \\
        IAE (PointNeXt) & \textbf{90.8} & \textbf{75.3} \\
        
        \bottomrule
    \end{tabular}
    \captionof{table}{\small{\textbf{Semantic Segmentation Results on S3DIS~\cite{armeni20163d} 6-Fold.} We show Overall Accuracy (OA) and mean Intersection over Union~(mIoU) across six folds.}}
    \label{tab:semseg}
\end{minipage}
\vspace{-4mm}
\end{table*}

\subsection{Pre-Training Setting}
\label{sec:setting}
As outlined in Section~\ref{sec:method}, we propose a general framework for Implicit Autoencoder. The IAE consists of an encoder part that maps a given point cloud into a high-dimensional latent code, and a decoder part that fits an implicit function to the latent code and query point to output the implicit function value. Our framework can easily incorporate any backbone architecture for point cloud understanding into the encoder part. In order to ensure a fair comparison, we adopt the same backbone architecture as previous approaches.

The decoder part of the IAE fits an implicit function that takes both the latent code obtained from the encoder and the query point as input, and outputs the implicit function value at that point. We explore two different designs for the decoder part based on Occupancy Network-style~\cite{mescheder2019occupancy} and Convolutional Occupancy Network-style~\cite{peng2020convolutional}. The pretrained models used for downstream tasks in our experiments are based on the Convolutional Occupancy Network-style decoder. Further implementation details are provided in the supplementary material.

\footnotetext{DepthContrast used a slightly larger model than Votenet. For a fair comparison, we reproduce DepthContrast with the original Votenet model.} 
\subsection{Downstream Tasks}
\label{sec:tasks}
After pre-training IAE on ShapeNet~\cite{chang2015shapenet} and ScanNet~\cite{dai2017scannet}, we discard the decoder in pre-training and append different network heads onto the pre-trained encoder for different downstream tasks, at both the object level and the scene level.
\vspace{-10pt}
\subsubsection{Shape Classification}
\label{sec:tasks:shape}
We assess the quality of object-level feature learning on ModelNet40~\cite{wu20153d} and ScanObjectNN ~\cite{uy2019revisiting} datasets. ModelNet40 consists of 9,832 training objects and 2,468 testing objects across 40 different categories. We additionally pre-process the dataset following the procedure developed by Qi et al~\cite{qi2017pointnet}. ScanObjectNN contains approximately 15,000 objects from 15 categories, which are scanned from real-world indoor scenes with cluttered backgrounds. We conduct experiments on three variants: OBJ-BG, OBJ-ONLY, and PBT50-RS. Details are provided in supplementary materials.

\noindent\textbf{Linear SVM Evaluation.} 
To evaluate the 3D representation learning ability of IAE, we first conduct linear classification experiments on ModelNet40~\cite{wu20153d} and ScanObjectNN~\cite{uy2019revisiting} after pre-training on ShapeNet. We use our frozen pre-trained encoder to extract features from 1,024 points sampled from each 3D shape and train a linear SVM classifier on top of them. The classification accuracies are reported in Table~\ref{tab:cls} LINEAR section. To establish a fair comparison with previous approaches, we construct two different encoders, DGCNN~\cite{wang2019dynamic} and M2AE~\cite{zhang2022point}. Under M2AE backbone, IAE achieves 93.2\%/84.4\% classification accuracy on the ModelNet40/ScanObjectNN testing split, representing a 0.3\%/1.3\% absolute improvement from the state-of-the-art method. Furthermore, we highlight that IAE is capable of surpassing state-of-the-art performance even without high-quality water-tight mesh labels. In the second-to-last row of Table~\ref{tab:cls} LINEAR section, we present the results when the ground-truth implicit representation is directly generated from the dense point cloud on ShapeNet using the nearest neighbor strategy. The 1.2\% absolute improvement on ScanObjectNN indicates that even though point clouds are discrete samples of the true 3D geometry, converting point clouds to the implicit format is an effective way of alleviating sampling variations.

\noindent\textbf{Supervised Fine-Tuning.} 
In addition to fitting an SVM in the latent space, we fine-tune the encoder pre-trained on ShapeNet using ModelNet40~\cite{wu20153d} and ScanObjectNN~\cite{uy2019revisiting} labels. We again adopt both DGCNN~\cite{wang2019dynamic} and M2AE~\cite{zhang2022point} architecture to construct the encoder network for a fair comparison, whose outputs are utilized as class predictions in this downstream task. As shown in Table~\ref{tab:cls} FULL section, IAE shows consistent improvements and achieves state-of-the-art performance with 94.3\%/88.2\% classification accuracy on ModelNet40/ScanObjectNN respectively. IAE also surpasses previous state-of-the-art with 94.2\%/88.0\% classification accuracy when the ground-truth implicit function is computed without meshes. 
\vspace{-10pt}
\begin{table}[t]
    \setlength\tabcolsep{10pt}
    \centering
    \small
    \begin{tabular}{lcc}
        \toprule
        Task & Pre-train & Acc/$\text{AP}_{25}$ \\
        \midrule
        \multirow{2}{*}{Object detection} & ScanNet & \textbf{67.6} \\
        & ShapeNet & 67.1 \\
        \midrule
        \multirow{2}{*}{MN40 Linear} & ScanNet & 91.3\% \\
        & ShapeNet & \textbf{92.1\%} \\
        \bottomrule
    \end{tabular}
    \captionof{table}{\small{\textbf{Cross-Domain Generalizability} between ShapeNet~\cite{chang2015shapenet} and ScanNet~\cite{dai2017scannet}. For the 3D object detection task, we report mAP at IoU=0.25 on the SUN RGB-D dataset~\cite{song2015sun}. For ModelNet40~\cite{wu20153d} linear evaluation, we report the classification accuracy.}}
    \label{tab:cross}
\vspace{-10pt}
\end{table}

\subsubsection{Indoor 3D Object Detection}
By leveraging the convolutional occupancy network~\cite{peng2020convolutional}, IAE demonstrates a robust ability to handle complex 3D scenes. We build the encoder with the VoteNet~\cite{qi2019deep} architecture and fine-tune the weights using ScanNet~\cite{dai2017scannet} detection labels after self-supervised pre-training on the same dataset. As shown in Table~\ref{tab:detection}, IAE outperforms random weight initialization with 18.8\% and 4.9\% relative improvements in mAP@0.5 and mAP@0.25, respectively. In addition to ScanNet, we further evaluate IAE on the more challenging SUN RGB-D dataset~\cite{song2015sun} and observe similar performance improvements.

Recently, Wang et al.~\cite{wang2022cagroup3d} propose the CAGroup3D model, which outperforms the commonly used VoteNet~\cite{qi2019deep} by a significant margin. Table~\ref{tab:detection} demonstrates that incorporating CAGroup3D as the IAE encoder further boosts the performance over the primitive CAGroup3D model.

\vspace{-2mm}
\subsubsection{Indoor 3D Semantic Segmentation}
On the Stanford Large-Scale 3D Indoor Spaces (S3DIS) dataset~\cite{armeni20163d}, we assess how well IAE can transfer its knowledge in self-supervised pre-training to scene-level semantic segmentation. S3DIS consists of 3D point clouds collected from 6 different large-area indoor environments, annotated with per-point categorical labels. Following Qi et al.~\cite{qi2017pointnet}, we divide each room instance into 1m $\times$ 1m blocks. We randomly sub-sample 4,096 points from the original point cloud as the encoder input and use 6-fold cross-validation during fine-tuning. As shown in Table~\ref{tab:semseg}, IAE significantly improves baseline approaches when the encoder is constructed from the same DGCNN~\cite{wang2019dynamic} architecture, with an 0.8\% absolute improvement in overall accuracy and 2.2\% improvement in mIoU. When integrated with PointNeXt~\cite{qian2022pointnext}, the latest model design in the field, IAE is still able to outperform the from-scratch training.

\begin{figure}[t]
    \centering
    \includegraphics[width=\columnwidth]{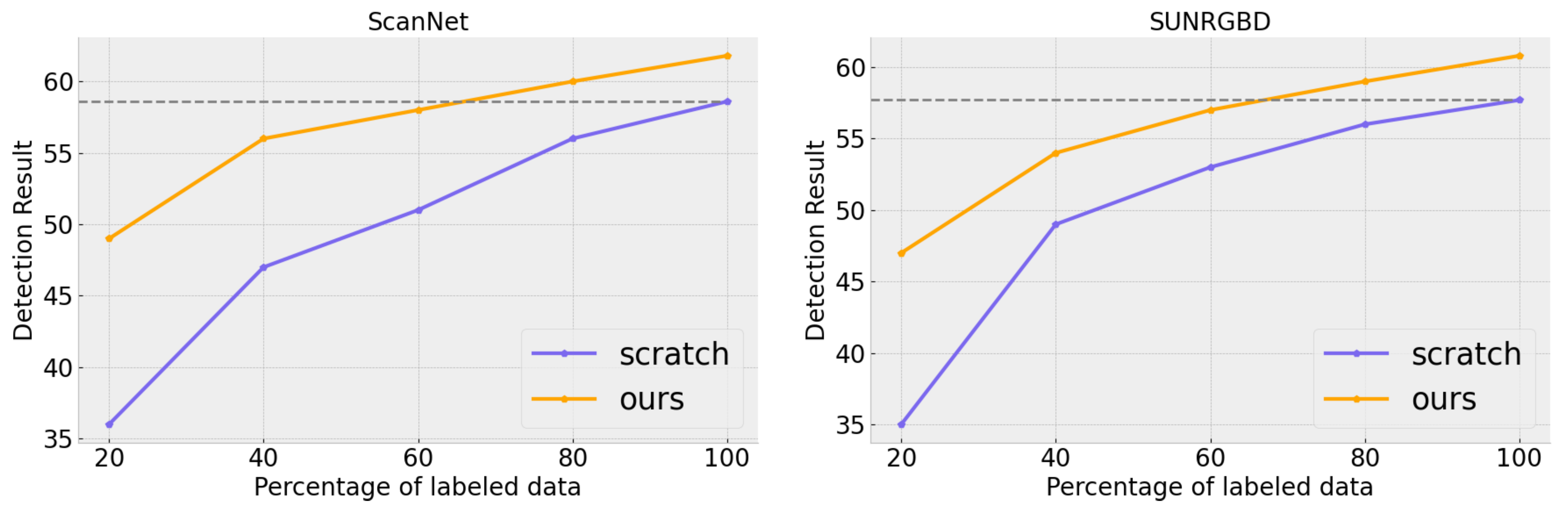}
    \caption{\textbf{Label efficiency training.} We pre-train our model on ScanNet and then fine-tune on ScanNet and SUN RGB-D separately. During fine-tuning, different percentages of labeled data are used. Our pre-training model outperforms training from scratch and achieves nearly the same result with only 60\% labeled data.}
    \label{fig:effi}
    \vspace{-10pt}
\end{figure}

\subsubsection{Cross-Domain Generalizability}
Up to this point, our discussion on scene-level fine-tuning assumes the pre-training process uses scene-level data as well. Whether synthetic object-level pre-training helps real-world scene-level fine-tuning remains an open debate in 3D vision. Xie et al.~\cite{xie2020pointcontrast} suggest pre-training on ShapeNet~\cite{chang2015shapenet} may have a detrimental effect on ScanNet~\cite{dai2017scannet} fine-tuning. In contrast, Huang et al.~\cite{huang2021spatio} demonstrate that with a sufficiently powerful encoder, object-level pre-training is actually beneficial regardless of whether the downstream task is at the object level or the scene level. As shown in Table~\ref{tab:cross}, with ShapeNet pre-training, IAE achieves a mAP@0.25 of 67.1 on the SUN RGB-D dataset~\cite{song2015sun} for object detection. Albeit not as good as scene-level pre-training, object-level pre-training demonstrates a noticeable improvement over training-from-scratch (66.8 mAP@0.25). On ModelNet40~\cite{wu20153d}, we use linear evaluation to determine the effect on scene-level classification. While it is meaningless to fit an SVM using an encoder with randomly initialized weights, Table~\ref{tab:cross} demonstrates that pre-training the IAE encoder on both ScanNet and ShapeNet leads to state-of-the-art classification accuracy under DGCNN~\cite{wang2019dynamic} backbone ({\em c.f.}, Table~\ref{tab:cls}). Empirically, we present additional evidence to support Huang et al.'s claim in the open debate.
\begin{table*}[t]
\footnotesize
\begin{minipage}[b]{.33\linewidth}
    \centering
    \begin{tabular}{l|cc}
        \toprule
         & Method & MN40 \\
        \midrule
        \multirow{3}{*}{EAE} &  FoldingNet & 90.1\% \\
        & OcCo & 89.7\%  \\
        & Snowflake &  89.9\% \\
        \midrule
        \multirow{2}{*}{IAE} & OccNet & 91.5\% \\
        & ConvONet &  \textbf{92.1\%} \\
        \bottomrule
    \end{tabular}
    \caption*{(a) Different decoders}
\end{minipage}
\begin{minipage}[b]{.33\linewidth}
    \centering
    \begin{tabular}{l|cc}
    \toprule
     & Method & SUN RGB-D \\
        \midrule
        \multirow{3}{*}{EAE} &  FoldingNet & 58.2 \\
        & OcCo & 58.4  \\
        & Snowflake & 58.1 \\
        \midrule
        IAE & ConvOccNet &  \textbf{60.4} \\
        \bottomrule
    \end{tabular}
    \caption*{(b) Different decoders on real data task}
\end{minipage}
\begin{minipage}[b]{.33\linewidth}
    \centering
    \begin{tabular}{l|cc}
        \toprule 
         & Function & MN40 \\
        \midrule
       EAE & PC & 90.1\% \\
        \midrule
        \multirow{4}{*}{IAE} & Occ & 91.3\% \\
        & UDF & 91.7\% \\
        & SDF & \textbf{92.1\%} \\
        \bottomrule
    \end{tabular}
    \caption*{(c) Different implicit functions}
\end{minipage}
\caption{\small{(a) \textbf{Ablation Study on Different Decoder Models} on shape-level ModelNet40~\cite{wu20153d} linear evaluation. Our implicit autoencoders (IAE) consistently outperform their explicit counterparts (EAE).  (b) \textbf{Ablation Study on Different Decoder Models} on real data task, mAP at IoU=0.25 on SUN RGB-D~\cite{song2015sun} detection. Our approach using the convolutional occupancy network shows consistent improvements. (c) \textbf{Ablation Study on Implicit Functions.} For explicit representation, we use FoldingNet~\cite{yang2018foldingnet} as the decoder and point cloud (PC) at the output. Three implicit representations show consistent improvements over the explicit representation.}}
\vspace{-10pt}
\label{tab:ablation}
\end{table*}
\subsubsection{Label Efficiency Training}
Pre-training helps models to be fine-tuned with a small amount of labeled data. We study the labeling efficiency of our model on 3D object detection by varying the portion of supervised training data. Results can be found in Figure~\ref{fig:effi}.
We take the VoteNet~\cite{qi2019deep} as the training network and use 20\%, 40\%, 60\%, and 80\% of the training data from ScanNet and SUN RGB-D datasets. We observe that our pre-training method gives larger gains when the labeled data is less. With only ~60\% training data on ScanNet/SUN RGB-D, our model can get similar performance compared with using all training data from scratch. This suggests our pre-training can help the downstream task to obtain better results with less data.

\begin{figure}[t]
    \centering
    \includegraphics[width=1.\columnwidth]{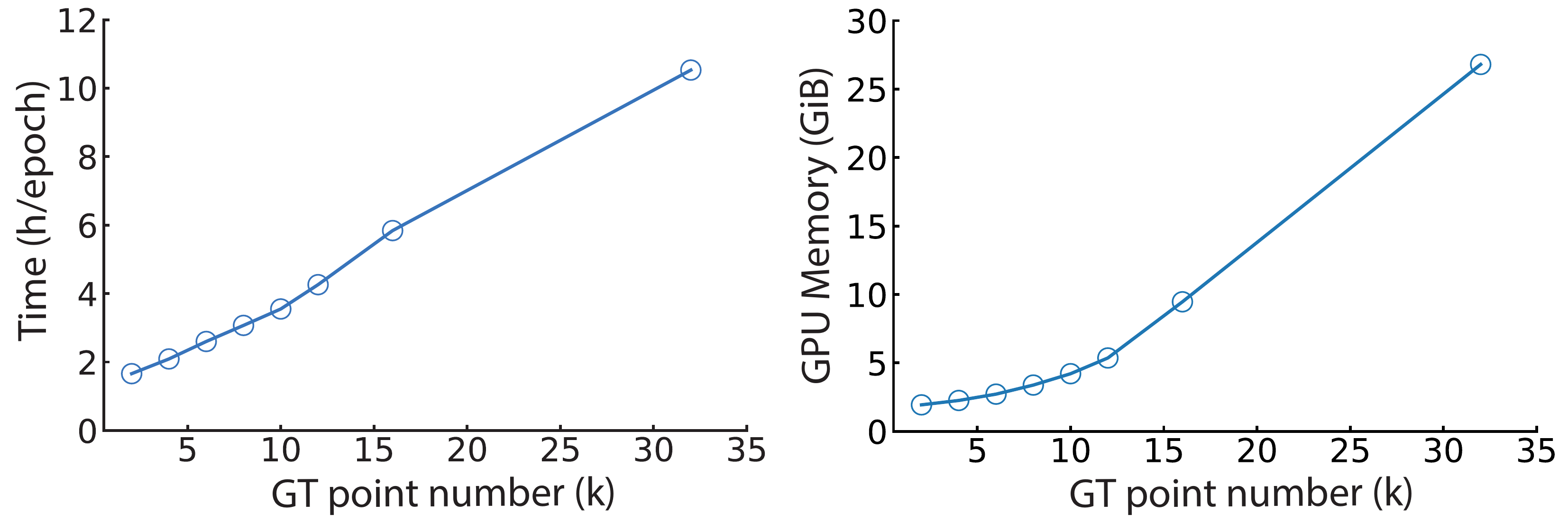}
    \caption{\small{\textbf{Computation time and GPU memory utilization of explicit autoencoder.} We use the standard explicit autoencoder framework, OcCo~\cite{wang2021unsupervised}, and take DGCNN~\cite{wang2019dynamic} as the encoder backbone.} We use Chamfer Distance as the loss function.}
    \label{fig:chamfer}
\end{figure}


\subsection{Ablation Study}
\label{sec:ablation}
We proceed with an ablation study to understand the different design choices we made in our experiments.

\noindent\textbf{Explicit vs. Implicit Decoders.} 
Under otherwise identical settings, we conduct two sets of parallel experiments to compare conventional explicit decoders with our proposed implicit decoders. Specifically, we choose three state-of-the-art explicit models, including FoldingNet~\cite{yang2018foldingnet}, OcCo~\cite{wang2020unsupervised}, and SnowflakeNet~\cite{xiang2021snowflakenet}, and compare them with two different implicit models, including Occupancy Network~\cite{mescheder2019occupancy}, and Convolutional Occupancy Network~\cite{peng2020convolutional}. We pre-train all models on ShapeNet~\cite{chang2015shapenet} and report the linear SVM classification accuracies on ModelNet40~\cite{wu20153d} under DGCNN~\cite{wang2019dynamic} backbone. As shown in Table~\ref{tab:ablation}~(a), implicit decoders consistently improve all explicit methods. Furthermore, we performed a comparison on real data by pre-training the models on ScanNet and reporting the SUN RGB-D detection result. As shown in Table~\ref{tab:ablation}~(b), the implicit decoder outperforms the explicit methods.

\noindent\textbf{Choice of Implicit Representations.}
Among the signed distance function~\cite{michalkiewicz2019implicit, park2019deepsdf}, unsigned distance function~\cite{chibane2020neural}, and occupancy grid~\cite{chen2019learning, mescheder2019occupancy, peng2020convolutional}, what is the implicit representation of choice in self-supervised representation learning? As shown in Table~\ref{tab:ablation}~(c), while they all outperform the explicit representation, the signed distance function demonstrates the highest linear evaluation accuracy on ModelNet40~\cite{wu20153d} classification with ShapeNet~\cite{chang2015shapenet} pre-training. When ground-truth signed distance functions are unavailable due to the lack of high-quality meshes, the unsigned distance function becomes the second-best choice.

\begin{figure}[t]
    \centering
    \begin{overpic}[width=\linewidth]{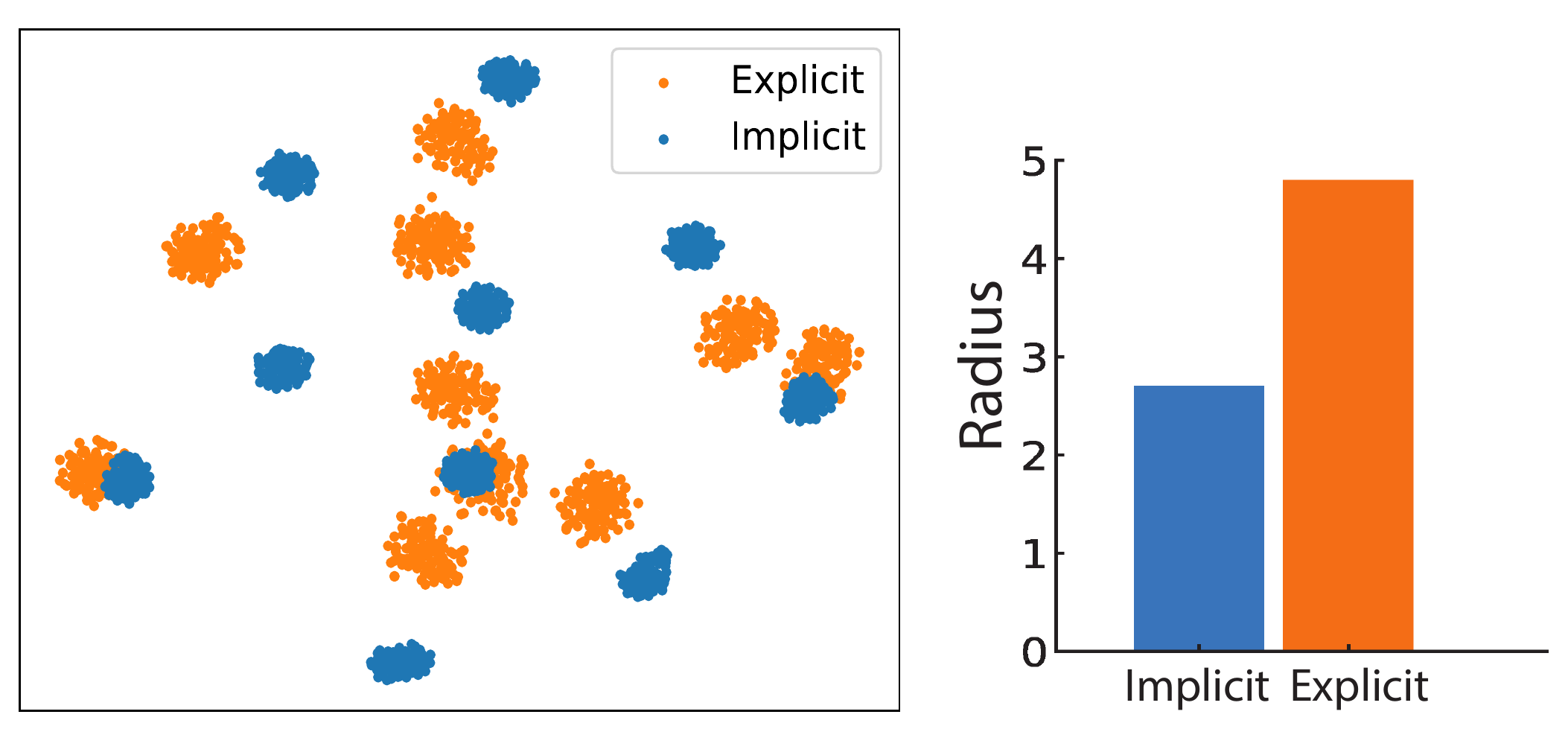}
    \put(27,-3.5){(a)}
    \put(80,-3.5){(b)}
    \end{overpic}
    \vspace{1pt}
    \caption{\small{\textbf{Experiment analysis on Explicit and Implicit Autoencoders.} \textbf{(a)} T-SNE visualization for encoder latent codes of IAE (blue) and EAE (orange). The IAE clusters are noticeably smaller. \textbf{(b)} The average cluster radius for IAE and EAE in (a).}}
    \label{fig:analysis}
    \vspace{-10pt}
\end{figure}

\subsection{Analysis}
\label{sec:analysis}
In this section, we show more experiment analysis between explicit autoencoders and implicit autoencoders.

\noindent\textbf{Computational Cost for Loss Function.}
As mentioned in Section~\ref{sec:challenge}, existing explicit autoencoders lack the ability to utilize dense point clouds from open 3D datasets, but sub-sampling inevitably introduces information loss and additional sampling variation. To verify this issue, we implemented the OcCo~\cite{wang2021unsupervised} framework and tested its computation time and GPU memory utilization as the ground truth point cloud density increased.
In Figure~\ref{fig:chamfer}, we observe that by applying Chamfer loss and training the model on a single Tesla V100 GPU with a batch size of 1, the training time increases to over 10.53 hours per epoch, and the GPU memory usage increases to 26.8 GiB as the number of points for the ground-truth point cloud increased to 32k. Since pre-training usually takes at least 200 epochs, this cost is unacceptable.
In comparison, IAE only takes 0.3 hours per epoch with 6 GiB GPU memory.

\noindent\textbf{Sampling Variation Problem.}
We conduct additional experiments to gain insights into the internal mechanism of IAE. Specifically, we pre-train both IAE and an explicit autoencoder, as discussed in Section~\ref{sec:tasks:shape}, on the ShapeNet dataset~\cite{chang2015shapenet}. Next, we randomly sample 100 different point clouds from 10 characteristic shapes and use the pre-trained encoders to predict latent codes.
We use T-SNE~\cite{van2008visualizing} to obtain embeddings of the latent codes in $\mathbb{R}^2$. As visualized in Figure~\ref{fig:analysis}~(a), both the implicit and explicit models are successful in mapping the shapes into 10 distinctive clusters. However, the implicit clusters are noticeably smaller in size, with an average radius of only 56.3\% of the explicit clusters (Figure~\ref{fig:analysis}~(b)). This result supports our claim that IAE is less sensitive to sampling variations and is able to learn more generalizable features about the true 3D geometry. 
\section{Analysis of \ouralg under a Linear AE model}
\label{Section:Motivation}

We proceed to gain more insights of \ouralg under a linear AE model. 


Specifically, suppose we have $N>n$ points $\bs{x}_i\in R^n, 1\leq i \leq N$. Without losing generality, we assume $\bs{x}_i$ lies on a low-dimensional linear space $\set{L}$ of dimension $m < n$. These data points are used to model the underlying 3D models. Now let us perturb each point $\bs{x}_i' = \bs{x}_i + \epsilon_i$, where $\epsilon_i$ is used to model sampling variations. We assume $\epsilon_i\in \set{L}^{\perp}$, meaning they encode variations that are orthogonal to variations among the underlying 3D models. Denote $X:=(\bs{x}_1,\cdots, \bs{x}_N)$ and $X':=(\bs{x}_1',\cdots, \bs{x}_{N}')$.

We consider two linear autoencoding models. The first one, which is analogous to \ouralg, takes $\bs{x}_i'$ as input and seeks to reconstruct $\bs{x}_i$:
\begin{align*}
A^{\star},Q^{\star} &= \underset{A', Q'\in \R^{n\times m}}{\textup{argmin}} \ \sum\limits_{k=1}^{N} \|A'{Q'}^T\bs{x}_{k}'-\bs{x}_k\|^2 \quad \\
s.t. \ {Q'}^TQ'& = I_m, \ Q'\in \set{C}(X')
\end{align*}
Here $Q'$ is the encoder, and $A'$ is the decoder. $\set{C}(X')$ denotes the column space of matrix $X'$. The constraints $Q'^T Q' = I_m$ and $Q'\in \set{C}(X')$ ensure that the encoder-decoder pair is unique up to a unitary transformation in $O(m)$.  
Below we show that $Q^{\star}$ is independent of $\epsilon_k$. 
\begin{prop}
Let $Q\in \R^{n\times m}$ collect the top-$m$ eigenvectors of the convariance matrix $C = \sum\limits_{k=1}^{N}\bs{x}_k\bs{x}_{k}'$. Then 
under the assumption that $\epsilon_k\in \set{L}^{\perp},1\leq k \leq N$,
$Q^{\star} = Q.$
\label{Prop:Derivative:Reformulated:AE}
\end{prop}
Prop.~\ref{Prop:Derivative:Reformulated:AE} indicates that $Q^{\star}$ does not encode sampling variations. 

Now consider the second model where we force the encoder-decoder pair to reconstruct the original inputs, which is analogous to standard autoencoding:
\begin{align*}
{\hat{A}^{\star}, \hat{Q}^{\star}} &= \underset{A',Q'\in \R^{n\times m}}{\textup{argmin}} \ \sum\limits_{k=1}^{N} \|A'{Q'}^T\bs{x}_k'-\bs{x}_k'\|^2_{\mathcal{F}} \quad \\
s.t. \ {Q'}^TQ' &= I_m, \ Q'\in \set{C}(X')
\end{align*}
In this case, $\hat{A}^{\star}=\hat{Q}^{\star}$, and both of them are given by the top $m$ eigenvectors of $C' = \sum\bs{x}_k'{\bs{x}_{k}'}^T$. 

To quantitatively compare encoders $\hat{Q}^{\star}$ and $Q$, we need the following definition.
\begin{defn}
Consider two unitary matrices $Q_1, Q_2 \in \R^{n\times m}$ where $Q_i^TQ_i = I_m$. We define the deviation between them as 
\begin{align*}
\mathcal{D}(Q_1, Q_2) &:= Q_1 -Q_2 R^{\star}, \\ \qquad R^{\star} &= \underset{R\in O(m)}{\textup{argmin}}\|Q_1 - Q_2R\|_{\mathcal{F}}^2
\label{Eq:Matrix:Difference}    
\end{align*}
\label{Defn:Matrix:Difference}
\end{defn}
The following proposition specifies the derivatives between $\hat{Q}^{\star}$ and $\epsilon_{k}$.
\begin{prop} 
Under the assumption that $\epsilon_k\in \set{L}^{\perp},1\leq k \leq N$, we have
\begin{equation}
\frac{\partial \mathcal{D}(\hat{Q}^{\star},Q)}{\partial \epsilon_{ki}} = (I_n - QQ^T)(\bs{e}_i\bs{x}_k^T)Q\Lambda^{+}
\label{Eq:Derivative:Q:Hat}
\end{equation}
where $\Lambda =\textup{diag}(\lambda_1,\cdots, \lambda_m)$ is a diagonal matrix that collects the top eigenvalues of $C$ that correspond to $Q$. $\bs{e}_k$ is the $k$-th standard basis vector.
\label{Prop:Derivative:Standard:AE}
\end{prop}
In other word, $\hat{Q}^{\star}$ is sensitive to $\epsilon_k$. Therefore, it encodes sampling variations. This theoretical analysis under a simplified setting suggests that \ouralg may potentially learn more robust representations.
\section{Conclusion}
\label{Section:Conclusions}
We present Implicit AutoEncoder (IAE), a simple yet effective model for point-cloud self-supervised representation learning. Unlike conventional point-cloud autoencoders, IAE exploits the implicit representation as the output of the decoder. IAE prevents latent space learning from being distracted by sampling variations and encourages the encoder to capture generalizable features from the true 3D geometry. Extensive experiments demonstrate that IAE achieves considerable improvements over a wide range of downstream tasks, including shape classification, linear evaluation, object detection, and indoor semantic segmentation. In the future, we plan to extend IAE to support not only hand-crafted but also trainable implicit representations that can be jointly learned with the autoencoder pipeline. 

\noindent\textbf{Acknowledgement.} Part of this work was initiated when Siming Yan was a summer research intern at Wormpex AI Research. Qixing Huang would like to acknowledge NSF
IIS-2047677, HDR-1934932, and CCF-2019844.



{\small
\bibliographystyle{ieee_fullname}
\bibliography{egbib}

\begin{thebibliography}{10}\itemsep=-1pt

\bibitem{achlioptas2018learning}
Panos Achlioptas, Olga Diamanti, Ioannis Mitliagkas, and Leonidas Guibas.
\newblock Learning representations and generative models for 3d point clouds.
\newblock In {\em International conference on machine learning}, pages 40--49.
  PMLR, 2018.

\bibitem{afham2022crosspoint}
Mohamed Afham, Isuru Dissanayake, Dinithi Dissanayake, Amaya Dharmasiri,
  Kanchana Thilakarathna, and Ranga Rodrigo.
\newblock Crosspoint: Self-supervised cross-modal contrastive learning for 3d
  point cloud understanding.
\newblock In {\em Proceedings of the IEEE/CVF Conference on Computer Vision and
  Pattern Recognition}, pages 9902--9912, 2022.

\bibitem{armeni20163d}
Iro Armeni, Ozan Sener, Amir~R Zamir, Helen Jiang, Ioannis Brilakis, Martin
  Fischer, and Silvio Savarese.
\newblock 3d semantic parsing of large-scale indoor spaces.
\newblock In {\em Proceedings of the IEEE Conference on Computer Vision and
  Pattern Recognition}, pages 1534--1543, 2016.

\bibitem{bengio2013representation}
Yoshua Bengio, Aaron Courville, and Pascal Vincent.
\newblock Representation learning: A review and new perspectives.
\newblock {\em IEEE transactions on pattern analysis and machine intelligence},
  35(8):1798--1828, 2013.

\bibitem{chang2015shapenet}
Angel~X Chang, Thomas Funkhouser, Leonidas Guibas, Pat Hanrahan, Qixing Huang,
  Zimo Li, Silvio Savarese, Manolis Savva, Shuran Song, Hao Su, et~al.
\newblock Shapenet: An information-rich 3d model repository.
\newblock {\em arXiv preprint arXiv:1512.03012}, 2015.

\bibitem{chen2020simple}
Ting Chen, Simon Kornblith, Mohammad Norouzi, and Geoffrey Hinton.
\newblock A simple framework for contrastive learning of visual
  representations.
\newblock In {\em International conference on machine learning}, pages
  1597--1607. PMLR, 2020.

\bibitem{chen2021shape}
Ye Chen, Jinxian Liu, Bingbing Ni, Hang Wang, Jiancheng Yang, Ning Liu, Teng
  Li, and Qi Tian.
\newblock Shape self-correction for unsupervised point cloud understanding.
\newblock In {\em Proceedings of the IEEE/CVF International Conference on
  Computer Vision}, pages 8382--8391, 2021.

\bibitem{chen2019learning}
Zhiqin Chen and Hao Zhang.
\newblock Learning implicit fields for generative shape modeling.
\newblock In {\em Proceedings of the IEEE/CVF Conference on Computer Vision and
  Pattern Recognition}, pages 5939--5948, 2019.

\bibitem{cheng2021back}
Bowen Cheng, Lu Sheng, Shaoshuai Shi, Ming Yang, and Dong Xu.
\newblock Back-tracing representative points for voting-based 3d object
  detection in point clouds.
\newblock In {\em Proceedings of the IEEE/CVF Conference on Computer Vision and
  Pattern Recognition}, pages 8963--8972, 2021.

\bibitem{chibane2020neural}
Julian Chibane, Gerard Pons-Moll, et~al.
\newblock Neural unsigned distance fields for implicit function learning.
\newblock {\em Advances in Neural Information Processing Systems},
  33:21638--21652, 2020.

\bibitem{dai2017scannet}
Angela Dai, Angel~X Chang, Manolis Savva, Maciej Halber, Thomas Funkhouser, and
  Matthias Nie{\ss}ner.
\newblock Scannet: Richly-annotated 3d reconstructions of indoor scenes.
\newblock In {\em Proceedings of the IEEE conference on computer vision and
  pattern recognition}, pages 5828--5839, 2017.

\bibitem{doersch2015unsupervised}
Carl Doersch, Abhinav Gupta, and Alexei~A Efros.
\newblock Unsupervised visual representation learning by context prediction.
\newblock In {\em Proceedings of the IEEE international conference on computer
  vision}, pages 1422--1430, 2015.

\bibitem{eckart2021self}
Benjamin Eckart, Wentao Yuan, Chao Liu, and Jan Kautz.
\newblock Self-supervised learning on 3d point clouds by learning discrete
  generative models.
\newblock In {\em Proceedings of the IEEE/CVF Conference on Computer Vision and
  Pattern Recognition}, pages 8248--8257, 2021.

\bibitem{engelmann20203d}
Francis Engelmann, Martin Bokeloh, Alireza Fathi, Bastian Leibe, and Matthias
  Nie{\ss}ner.
\newblock 3d-mpa: Multi-proposal aggregation for 3d semantic instance
  segmentation.
\newblock In {\em Proceedings of the IEEE/CVF conference on computer vision and
  pattern recognition}, pages 9031--9040, 2020.

\bibitem{erhan2010does}
Dumitru Erhan, Aaron Courville, Yoshua Bengio, and Pascal Vincent.
\newblock Why does unsupervised pre-training help deep learning?
\newblock In {\em Proceedings of the thirteenth international conference on
  artificial intelligence and statistics}, pages 201--208. JMLR Workshop and
  Conference Proceedings, 2010.

\bibitem{fan2017point}
Haoqiang Fan, Hao Su, and Leonidas~J Guibas.
\newblock A point set generation network for 3d object reconstruction from a
  single image.
\newblock In {\em Proceedings of the IEEE conference on computer vision and
  pattern recognition}, pages 605--613, 2017.

\bibitem{geiger2012we}
Andreas Geiger, Philip Lenz, and Raquel Urtasun.
\newblock Are we ready for autonomous driving? the kitti vision benchmark
  suite.
\newblock In {\em 2012 IEEE conference on computer vision and pattern
  recognition}, pages 3354--3361. IEEE, 2012.

\bibitem{genova2020local}
Kyle Genova, Forrester Cole, Avneesh Sud, Aaron Sarna, and Thomas Funkhouser.
\newblock Local deep implicit functions for 3d shape.
\newblock In {\em Proceedings of the IEEE/CVF Conference on Computer Vision and
  Pattern Recognition}, pages 4857--4866, 2020.

\bibitem{genova2019learning}
Kyle Genova, Forrester Cole, Daniel Vlasic, Aaron Sarna, William~T Freeman, and
  Thomas Funkhouser.
\newblock Learning shape templates with structured implicit functions.
\newblock In {\em Proceedings of the IEEE/CVF International Conference on
  Computer Vision}, pages 7154--7164, 2019.

\bibitem{hassani2019unsupervised}
Kaveh Hassani and Mike Haley.
\newblock Unsupervised multi-task feature learning on point clouds.
\newblock In {\em Proceedings of the IEEE/CVF International Conference on
  Computer Vision}, pages 8160--8171, 2019.

\bibitem{he2021masked}
Kaiming He, Xinlei Chen, Saining Xie, Yanghao Li, Piotr Doll{\'a}r, and Ross
  Girshick.
\newblock Masked autoencoders are scalable vision learners.
\newblock {\em arXiv preprint arXiv:2111.06377}, 2021.

\bibitem{hu2020randla}
Qingyong Hu, Bo Yang, Linhai Xie, Stefano Rosa, Yulan Guo, Zhihua Wang, Niki
  Trigoni, and Andrew Markham.
\newblock Randla-net: Efficient semantic segmentation of large-scale point
  clouds.
\newblock In {\em Proceedings of the IEEE/CVF conference on computer vision and
  pattern recognition}, pages 11108--11117, 2020.

\bibitem{huang2021spatio}
Siyuan Huang, Yichen Xie, Song-Chun Zhu, and Yixin Zhu.
\newblock Spatio-temporal self-supervised representation learning for 3d point
  clouds.
\newblock In {\em Proceedings of the IEEE/CVF International Conference on
  Computer Vision}, pages 6535--6545, 2021.

\bibitem{li2021deepgcns}
Guohao Li, Matthias M{\"u}ller, Guocheng Qian, Itzel Carolina~Delgadillo Perez,
  Abdulellah Abualshour, Ali~Kassem Thabet, and Bernard Ghanem.
\newblock Deepgcns: Making gcns go as deep as cnns.
\newblock {\em IEEE transactions on pattern analysis and machine intelligence},
  2021.

\bibitem{li2018so}
Jiaxin Li, Ben~M Chen, and Gim~Hee Lee.
\newblock So-net: Self-organizing network for point cloud analysis.
\newblock In {\em Proceedings of the IEEE conference on computer vision and
  pattern recognition}, pages 9397--9406, 2018.

\bibitem{li2018pointcnn}
Yangyan Li, Rui Bu, Mingchao Sun, Wei Wu, Xinhan Di, and Baoquan Chen.
\newblock Pointcnn: Convolution on x-transformed points.
\newblock {\em Advances in neural information processing systems}, 31:820--830,
  2018.

\bibitem{liu2019learning}
Shichen Liu, Shunsuke Saito, Weikai Chen, and Hao Li.
\newblock Learning to infer implicit surfaces without 3d supervision.
\newblock {\em arXiv preprint arXiv:1911.00767}, 2019.

\bibitem{liu2020dist}
Shaohui Liu, Yinda Zhang, Songyou Peng, Boxin Shi, Marc Pollefeys, and Zhaopeng
  Cui.
\newblock Dist: Rendering deep implicit signed distance function with
  differentiable sphere tracing.
\newblock In {\em Proceedings of the IEEE/CVF Conference on Computer Vision and
  Pattern Recognition}, pages 2019--2028, 2020.

\bibitem{liu2021group}
Ze Liu, Zheng Zhang, Yue Cao, Han Hu, and Xin Tong.
\newblock Group-free 3d object detection via transformers.
\newblock In {\em Proceedings of the IEEE/CVF International Conference on
  Computer Vision}, pages 2949--2958, 2021.

\bibitem{ma2022rethinking}
Xu Ma, Can Qin, Haoxuan You, Haoxi Ran, and Yun Fu.
\newblock Rethinking network design and local geometry in point cloud: A simple
  residual mlp framework.
\newblock {\em arXiv preprint arXiv:2202.07123}, 2022.

\bibitem{masci2011stacked}
Jonathan Masci, Ueli Meier, Dan Cire{\c{s}}an, and J{\"u}rgen Schmidhuber.
\newblock Stacked convolutional auto-encoders for hierarchical feature
  extraction.
\newblock In {\em International conference on artificial neural networks},
  pages 52--59. Springer, 2011.

\bibitem{mescheder2019occupancy}
Lars Mescheder, Michael Oechsle, Michael Niemeyer, Sebastian Nowozin, and
  Andreas Geiger.
\newblock Occupancy networks: Learning 3d reconstruction in function space.
\newblock In {\em Proceedings of the IEEE/CVF Conference on Computer Vision and
  Pattern Recognition}, pages 4460--4470, 2019.

\bibitem{michalkiewicz2019implicit}
Mateusz Michalkiewicz, Jhony~K Pontes, Dominic Jack, Mahsa Baktashmotlagh, and
  Anders Eriksson.
\newblock Implicit surface representations as layers in neural networks.
\newblock In {\em Proceedings of the IEEE/CVF International Conference on
  Computer Vision}, pages 4743--4752, 2019.

\bibitem{misra2021end}
Ishan Misra, Rohit Girdhar, and Armand Joulin.
\newblock An end-to-end transformer model for 3d object detection.
\newblock In {\em Proceedings of the IEEE/CVF International Conference on
  Computer Vision}, pages 2906--2917, 2021.

\bibitem{niemeyer2019occupancy}
Michael Niemeyer, Lars Mescheder, Michael Oechsle, and Andreas Geiger.
\newblock Occupancy flow: 4d reconstruction by learning particle dynamics.
\newblock In {\em Proceedings of the IEEE/CVF International Conference on
  Computer Vision}, pages 5379--5389, 2019.

\bibitem{niemeyer2020differentiable}
Michael Niemeyer, Lars Mescheder, Michael Oechsle, and Andreas Geiger.
\newblock Differentiable volumetric rendering: Learning implicit 3d
  representations without 3d supervision.
\newblock In {\em Proceedings of the IEEE/CVF Conference on Computer Vision and
  Pattern Recognition}, pages 3504--3515, 2020.

\bibitem{oechsle2019texture}
Michael Oechsle, Lars Mescheder, Michael Niemeyer, Thilo Strauss, and Andreas
  Geiger.
\newblock Texture fields: Learning texture representations in function space.
\newblock In {\em Proceedings of the IEEE/CVF International Conference on
  Computer Vision}, pages 4531--4540, 2019.

\bibitem{pang2022masked}
Yatian Pang, Wenxiao Wang, Francis~EH Tay, Wei Liu, Yonghong Tian, and Li Yuan.
\newblock Masked autoencoders for point cloud self-supervised learning.
\newblock {\em arXiv preprint arXiv:2203.06604}, 2022.

\bibitem{park2019deepsdf}
Jeong~Joon Park, Peter Florence, Julian Straub, Richard Newcombe, and Steven
  Lovegrove.
\newblock Deepsdf: Learning continuous signed distance functions for shape
  representation.
\newblock In {\em Proceedings of the IEEE/CVF Conference on Computer Vision and
  Pattern Recognition}, pages 165--174, 2019.

\bibitem{paschalidou2020learning}
Despoina Paschalidou, Luc~Van Gool, and Andreas Geiger.
\newblock Learning unsupervised hierarchical part decomposition of 3d objects
  from a single rgb image.
\newblock In {\em Proceedings of the IEEE/CVF Conference on Computer Vision and
  Pattern Recognition}, pages 1060--1070, 2020.

\bibitem{pathak2016context}
Deepak Pathak, Philipp Krahenbuhl, Jeff Donahue, Trevor Darrell, and Alexei~A
  Efros.
\newblock Context encoders: Feature learning by inpainting.
\newblock In {\em Proceedings of the IEEE conference on computer vision and
  pattern recognition}, pages 2536--2544, 2016.

\bibitem{peng2020convolutional}
Songyou Peng, Michael Niemeyer, Lars Mescheder, Marc Pollefeys, and Andreas
  Geiger.
\newblock Convolutional occupancy networks.
\newblock In {\em Computer Vision--ECCV 2020: 16th European Conference,
  Glasgow, UK, August 23--28, 2020, Proceedings, Part III 16}, pages 523--540.
  Springer, 2020.

\bibitem{peterson2009k}
Leif~E Peterson.
\newblock K-nearest neighbor.
\newblock {\em Scholarpedia}, 4(2):1883, 2009.

\bibitem{poursaeed2020self}
Omid Poursaeed, Tianxing Jiang, Han Qiao, Nayun Xu, and Vladimir~G Kim.
\newblock Self-supervised learning of point clouds via orientation estimation.
\newblock In {\em 2020 International Conference on 3D Vision (3DV)}, pages
  1018--1028. IEEE, 2020.

\bibitem{qi2019deep}
Charles~R Qi, Or Litany, Kaiming He, and Leonidas~J Guibas.
\newblock Deep hough voting for 3d object detection in point clouds.
\newblock In {\em Proceedings of the IEEE/CVF International Conference on
  Computer Vision}, pages 9277--9286, 2019.

\bibitem{qi2017pointnet}
Charles~R Qi, Hao Su, Kaichun Mo, and Leonidas~J Guibas.
\newblock Pointnet: Deep learning on point sets for 3d classification and
  segmentation.
\newblock In {\em Proceedings of the IEEE conference on computer vision and
  pattern recognition}, pages 652--660, 2017.

\bibitem{qi2017pointnet++}
Charles~R Qi, Li Yi, Hao Su, and Leonidas~J Guibas.
\newblock Pointnet++: Deep hierarchical feature learning on point sets in a
  metric space.
\newblock {\em arXiv preprint arXiv:1706.02413}, 2017.

\bibitem{qian2022pointnext}
Guocheng Qian, Yuchen Li, Houwen Peng, Jinjie Mai, Hasan Abed Al~Kader Hammoud,
  Mohamed Elhoseiny, and Bernard Ghanem.
\newblock Pointnext: Revisiting pointnet++ with improved training and scaling
  strategies.
\newblock {\em arXiv preprint arXiv:2206.04670}, 2022.

\bibitem{qiu2021geometric}
Shi Qiu, Saeed Anwar, and Nick Barnes.
\newblock Geometric back-projection network for point cloud classification.
\newblock {\em IEEE Transactions on Multimedia}, 24:1943--1955, 2021.

\bibitem{qiu2021semantic}
Shi Qiu, Saeed Anwar, and Nick Barnes.
\newblock Semantic segmentation for real point cloud scenes via bilateral
  augmentation and adaptive fusion.
\newblock In {\em Proceedings of the IEEE/CVF Conference on Computer Vision and
  Pattern Recognition}, pages 1757--1767, 2021.

\bibitem{rao2021randomrooms}
Yongming Rao, Benlin Liu, Yi Wei, Jiwen Lu, Cho-Jui Hsieh, and Jie Zhou.
\newblock Randomrooms: Unsupervised pre-training from synthetic shapes and
  randomized layouts for 3d object detection.
\newblock In {\em Proceedings of the IEEE/CVF International Conference on
  Computer Vision}, pages 3283--3292, 2021.

\bibitem{rubner2000earth}
Yossi Rubner, Carlo Tomasi, and Leonidas~J Guibas.
\newblock The earth mover's distance as a metric for image retrieval.
\newblock {\em International journal of computer vision}, 40(2):99--121, 2000.

\bibitem{rukhovich2021fcaf3d}
Danila Rukhovich, Anna Vorontsova, and Anton Konushin.
\newblock Fcaf3d: Fully convolutional anchor-free 3d object detection.
\newblock {\em arXiv preprint arXiv:2112.00322}, 2021.

\bibitem{sauder2019self}
Jonathan Sauder and Bjarne Sievers.
\newblock Self-supervised deep learning on point clouds by reconstructing
  space.
\newblock {\em arXiv preprint arXiv:1901.08396}, 2019.

\bibitem{sharma2020parsenet}
Gopal Sharma, Difan Liu, Subhransu Maji, Evangelos Kalogerakis, Siddhartha
  Chaudhuri, and Radom{\'\i}r M{\v{e}}ch.
\newblock Parsenet: A parametric surface fitting network for 3d point clouds.
\newblock In {\em Computer Vision--ECCV 2020: 16th European Conference,
  Glasgow, UK, August 23--28, 2020, Proceedings, Part VII 16}, pages 261--276.
  Springer, 2020.

\bibitem{song2015sun}
Shuran Song, Samuel~P Lichtenberg, and Jianxiong Xiao.
\newblock Sun rgb-d: A rgb-d scene understanding benchmark suite.
\newblock In {\em Proceedings of the IEEE conference on computer vision and
  pattern recognition}, pages 567--576, 2015.

\bibitem{tang2022contrastive}
Liyao Tang, Yibing Zhan, Zhe Chen, Baosheng Yu, and Dacheng Tao.
\newblock Contrastive boundary learning for point cloud segmentation.
\newblock In {\em Proceedings of the IEEE/CVF Conference on Computer Vision and
  Pattern Recognition}, pages 8489--8499, 2022.

\bibitem{thomas2019kpconv}
Hugues Thomas, Charles~R Qi, Jean-Emmanuel Deschaud, Beatriz Marcotegui,
  Fran{\c{c}}ois Goulette, and Leonidas~J Guibas.
\newblock Kpconv: Flexible and deformable convolution for point clouds.
\newblock In {\em Proceedings of the IEEE/CVF International Conference on
  Computer Vision}, pages 6411--6420, 2019.

\bibitem{tschannen2018recent}
Michael Tschannen, Olivier Bachem, and Mario Lucic.
\newblock Recent advances in autoencoder-based representation learning.
\newblock {\em arXiv preprint arXiv:1812.05069}, 2018.

\bibitem{uy-scanobjectnn-iccv19}
Mikaela~Angelina Uy, Quang-Hieu Pham, Binh-Son Hua, Duc~Thanh Nguyen, and
  Sai-Kit Yeung.
\newblock Revisiting point cloud classification: A new benchmark dataset and
  classification model on real-world data.
\newblock In {\em International Conference on Computer Vision (ICCV)}, 2019.

\bibitem{uy2019revisiting}
Mikaela~Angelina Uy, Quang-Hieu Pham, Binh-Son Hua, Thanh Nguyen, and Sai-Kit
  Yeung.
\newblock Revisiting point cloud classification: A new benchmark dataset and
  classification model on real-world data.
\newblock In {\em Proceedings of the IEEE/CVF international conference on
  computer vision}, pages 1588--1597, 2019.

\bibitem{van2008visualizing}
Laurens Van~der Maaten and Geoffrey Hinton.
\newblock Visualizing data using t-sne.
\newblock {\em Journal of machine learning research}, 9(11), 2008.

\bibitem{vincent2008extracting}
Pascal Vincent, Hugo Larochelle, Yoshua Bengio, and Pierre-Antoine Manzagol.
\newblock Extracting and composing robust features with denoising autoencoders.
\newblock In {\em Proceedings of the 25th international conference on Machine
  learning}, pages 1096--1103, 2008.

\bibitem{wang2022cagroup3d}
Haiyang Wang, Lihe Ding, Shaocong Dong, Shaoshuai Shi, Aoxue Li, Jianan Li,
  Zhenguo Li, and Liwei Wang.
\newblock Cagroup3d: Class-aware grouping for 3d object detection on point
  clouds.
\newblock {\em arXiv preprint arXiv:2210.04264}, 2022.

\bibitem{wang2021unsupervised}
Hanchen Wang, Qi Liu, Xiangyu Yue, Joan Lasenby, and Matt~J Kusner.
\newblock Unsupervised point cloud pre-training via occlusion completion.
\newblock In {\em Proceedings of the IEEE/CVF international conference on
  computer vision}, pages 9782--9792, 2021.

\bibitem{wang2020unsupervised}
Peng-Shuai Wang, Yu-Qi Yang, Qian-Fang Zou, Zhirong Wu, Yang Liu, and Xin Tong.
\newblock Unsupervised 3d learning for shape analysis via multiresolution
  instance discrimination.
\newblock {\em ACM Trans. Graphic}, 2020.

\bibitem{wang2019dynamic}
Yue Wang, Yongbin Sun, Ziwei Liu, Sanjay~E Sarma, Michael~M Bronstein, and
  Justin~M Solomon.
\newblock Dynamic graph cnn for learning on point clouds.
\newblock {\em Acm Transactions On Graphics (tog)}, 38(5):1--12, 2019.

\bibitem{wu20153d}
Zhirong Wu, Shuran Song, Aditya Khosla, Fisher Yu, Linguang Zhang, Xiaoou Tang,
  and Jianxiong Xiao.
\newblock 3d shapenets: A deep representation for volumetric shapes.
\newblock In {\em Proceedings of the IEEE conference on computer vision and
  pattern recognition}, pages 1912--1920, 2015.

\bibitem{wu2018unsupervised}
Zhirong Wu, Yuanjun Xiong, Stella~X Yu, and Dahua Lin.
\newblock Unsupervised feature learning via non-parametric instance
  discrimination.
\newblock In {\em Proceedings of the IEEE conference on computer vision and
  pattern recognition}, pages 3733--3742, 2018.

\bibitem{xiang2021snowflakenet}
Peng Xiang, Xin Wen, Yu-Shen Liu, Yan-Pei Cao, Pengfei Wan, Wen Zheng, and
  Zhizhong Han.
\newblock Snowflakenet: Point cloud completion by snowflake point deconvolution
  with skip-transformer.
\newblock In {\em Proceedings of the IEEE/CVF International Conference on
  Computer Vision}, pages 5499--5509, 2021.

\bibitem{xie2020mlcvnet}
Qian Xie, Yu-Kun Lai, Jing Wu, Zhoutao Wang, Yiming Zhang, Kai Xu, and Jun
  Wang.
\newblock Mlcvnet: Multi-level context votenet for 3d object detection.
\newblock In {\em Proceedings of the IEEE/CVF conference on computer vision and
  pattern recognition}, pages 10447--10456, 2020.

\bibitem{xie2020pointcontrast}
Saining Xie, Jiatao Gu, Demi Guo, Charles~R Qi, Leonidas Guibas, and Or Litany.
\newblock Pointcontrast: Unsupervised pre-training for 3d point cloud
  understanding.
\newblock In {\em European Conference on Computer Vision}, pages 574--591.
  Springer, 2020.

\bibitem{yan2023learning}
Siming Yan, Chen Song, Youkang Kong, and Qixing Huang.
\newblock Learning from multi-view representation for point-cloud pre-training.
\newblock {\em arXiv preprint arXiv:2306.02558}, 2023.

\bibitem{yan20233d}
Siming Yan, Yuqi Yang, Yuxiao Guo, Hao Pan, Peng-shuai Wang, Xin Tong, Yang
  Liu, and Qixing Huang.
\newblock 3d feature prediction for masked-autoencoder-based point cloud
  pretraining.
\newblock {\em arXiv preprint arXiv:2304.06911}, 2023.

\bibitem{yan2021hpnet}
Siming Yan, Zhenpei Yang, Chongyang Ma, Haibin Huang, Etienne Vouga, and Qixing
  Huang.
\newblock Hpnet: Deep primitive segmentation using hybrid representations.
\newblock {\em arXiv preprint arXiv:2105.10620}, 2021.

\bibitem{yang2018foldingnet}
Yaoqing Yang, Chen Feng, Yiru Shen, and Dong Tian.
\newblock Foldingnet: Point cloud auto-encoder via deep grid deformation.
\newblock In {\em Proceedings of the IEEE Conference on Computer Vision and
  Pattern Recognition}, pages 206--215, 2018.

\bibitem{yi2016scalable}
Li Yi, Vladimir~G Kim, Duygu Ceylan, I-Chao Shen, Mengyan Yan, Hao Su, Cewu Lu,
  Qixing Huang, Alla Sheffer, and Leonidas Guibas.
\newblock A scalable active framework for region annotation in 3d shape
  collections.
\newblock {\em ACM Transactions on Graphics (ToG)}, 35(6):1--12, 2016.

\bibitem{yu2021point}
Xumin Yu, Lulu Tang, Yongming Rao, Tiejun Huang, Jie Zhou, and Jiwen Lu.
\newblock Point-bert: Pre-training 3d point cloud transformers with masked
  point modeling.
\newblock {\em arXiv preprint arXiv:2111.14819}, 2021.

\bibitem{zhang2022point}
Renrui Zhang, Ziyu Guo, Peng Gao, Rongyao Fang, Bin Zhao, Dong Wang, Yu Qiao,
  and Hongsheng Li.
\newblock Point-m2ae: multi-scale masked autoencoders for hierarchical point
  cloud pre-training.
\newblock {\em arXiv preprint arXiv:2205.14401}, 2022.

\bibitem{zhang2021self}
Zaiwei Zhang, Rohit Girdhar, Armand Joulin, and Ishan Misra.
\newblock Self-supervised pretraining of 3d features on any point-cloud.
\newblock {\em arXiv preprint arXiv:2101.02691}, 2021.

\bibitem{zhao2021point}
Hengshuang Zhao, Li Jiang, Jiaya Jia, Philip~HS Torr, and Vladlen Koltun.
\newblock Point transformer.
\newblock In {\em Proceedings of the IEEE/CVF International Conference on
  Computer Vision}, pages 16259--16268, 2021.

\bibitem{zhuang2021unsupervised}
Chengxu Zhuang, Siming Yan, Aran Nayebi, Martin Schrimpf, Michael~C Frank,
  James~J DiCarlo, and Daniel~LK Yamins.
\newblock Unsupervised neural network models of the ventral visual stream.
\newblock {\em Proceedings of the National Academy of Sciences}, 118(3), 2021.

\bibitem{zhuang2019self}
Chengxu Zhuang, Siming Yan, Aran Nayebi, and Daniel Yamins.
\newblock Self-supervised neural network models of higher visual cortex
  development.
\newblock In {\em 2019 Conference on Cognitive Computational Neuroscience},
  pages 566--569. CCN, 2019.

\end{thebibliography}
}

\end{document}